\pdfoutput=1

\documentclass[times,twocolumn,final,authoryear]{elsarticle}

\usepackage{hyperref}
\usepackage{footnote}

\usepackage{prletters}
\usepackage{framed,multirow}
\usepackage{amsmath}

\usepackage{amssymb}
\usepackage{latexsym}

\usepackage{url}
\usepackage{xcolor}
\definecolor{newcolor}{rgb}{.8,.349,.1}

\journal{Image and Vision Computing}

\begin{document}

\thispagestyle{empty}

%%%%%%%% START PAPER

\clearpage

\ifpreprint
  \setcounter{page}{1}
\else
  \setcounter{page}{1}
\fi

\begin{frontmatter}

\title{Ego-Object Discovery}

\author[1]{Marc \snm{Bola\~nos}}
\cortext[cor1]{Marc Bola\~nos: 
  Tel.: +34-669-648-301}
\ead{marc.bolanos@ub.edu}
\author[1,2]{Petia \snm{Radeva}}

\address[1]{Universitat de Barcelona, Gran Via de les Corts Catalanes, 585, Barcelona 08007, Spain}
\address[2]{Computer Vision Center, Building O Campus UAB, Bellaterra (Barcelona) 08193, Spain}

%\received{1 May 2013}
%\finalform{10 May 2013}
%\accepted{13 May 2013}
%\availableonline{15 May 2013}
%\communicated{S. Sarkar}

%========================================================%
%========================================================%
%				ABSTRACT
%========================================================%
\begin{abstract}
Lifelogging devices are spreading faster everyday. This growth can represent great benefits to develop methods for extraction of meaningful information about the user wearing the device and his/her environment. In this paper, we propose a semi-supervised strategy for easily discovering objects relevant to the person wearing a first-person camera. Given an egocentric video/images sequence acquired by the camera, our algorithm uses both the appearance extracted by means of a convolutional neural network and an object refill methodology that allows to discover objects even in case of small amount of object appearance in the collection of images. An SVM filtering strategy is applied to deal with the great part of the False Positive object candidates found by most of the state of the art object detectors. We validate our method on a new egocentric dataset of 4912 daily images acquired by 4 persons as well as on both PASCAL 2012 and MSRC datasets. We obtain for all of them results that largely outperform the state of the art approach. We make public both the EDUB dataset\footnotemark[1] and the algorithm code\footnotemark[2].
    % , and by nearly a 450\% on average. 
\end{abstract}
%\keywords{object discovery, object recognition, egocentric videos, lifelogging, CNN}

\end{frontmatter}

%========================================================%
%========================================================%
%				INTRODUCTION
%========================================================%
%========================================================%
\section{Introduction}

\footnotetext[1]{\href{https://www.dropbox.com/s/py8xhalqxz15co3/EDUB\%202015.zip?dl=0}{https://www.dropbox.com/s/py8xhalqxz15co3/EDUB\%202015.zip?dl=0}}
\footnotetext[2]{\href{https://github.com/MarcBS/Ego-Object\_Discovery/releases}{https://github.com/MarcBS/Ego-Object\_Discovery/releases}}
Ubiquitous computing is more present everyday in our lives, and with it lifelogging devices \citep{hodges2006sensecam,michael2013wearable} are increasing their popularity and spread. By using wearable cameras, we can acquire continuous data about the life of persons, and build applications that convert this huge amount of data into meaningful information about their lifestyle. Hence, wearable cameras offer an easy manner to acquire information about our daily life tasks, and extract information about our typical activities and habits \citep{betancourtevolution} from an egocentric (or first-person) point of view. 
\begin{figure}[!hb]
  \centering
  \fbox{\includegraphics[width=\columnwidth]{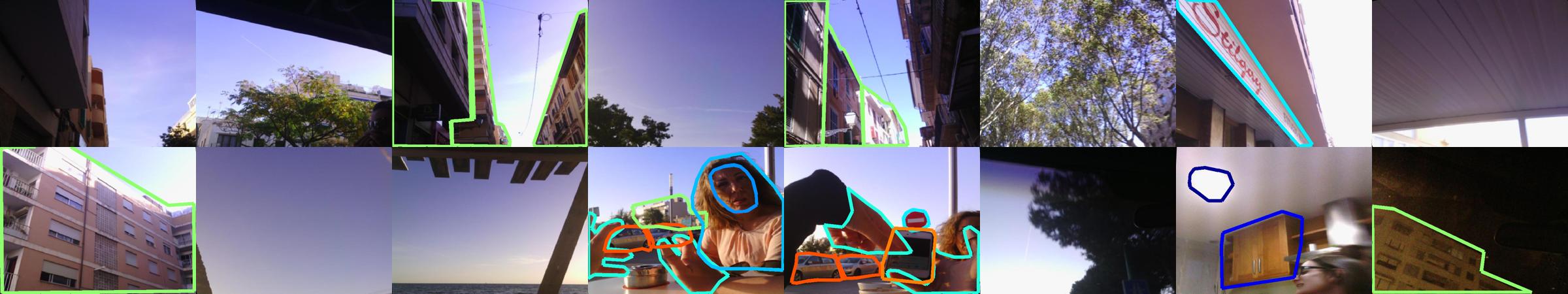}}
  \fbox{\includegraphics[width=\columnwidth]{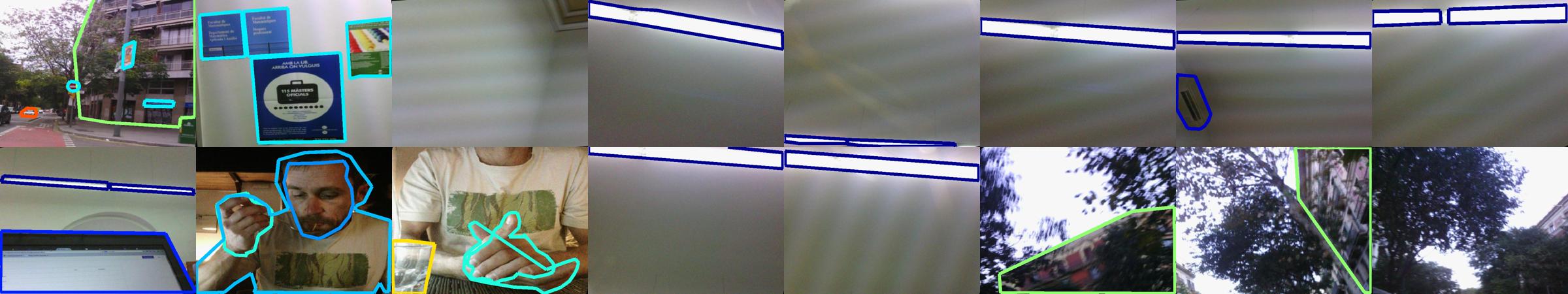}}
  \fbox{\includegraphics[width=\columnwidth]{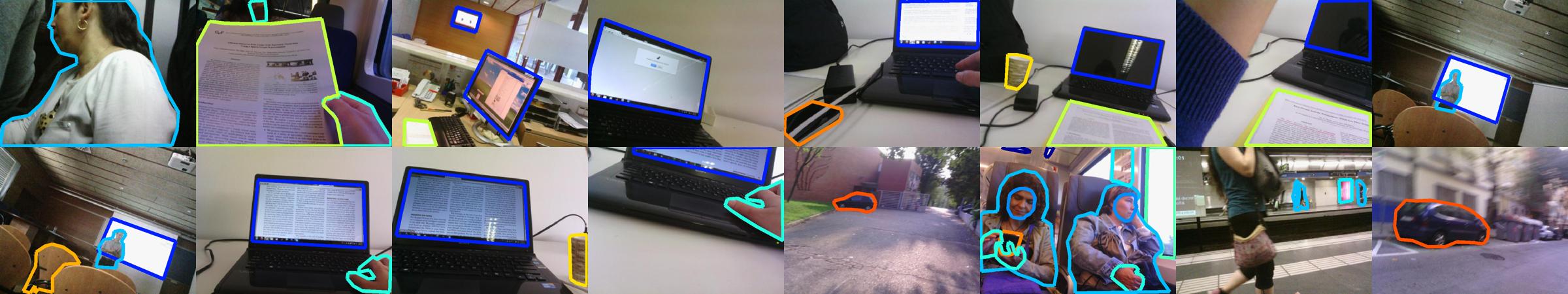}}
  \caption{Lifelogging sets from 3 users (each 2 row correspond to a different user). Note how objects help to discriminate different environments. The annotated objects are to be discovered by the object discovery algorithm.}
  \label{fig:objects_person}
\end{figure}
For example, Fig. \ref{fig:objects_person} shows datasets acquired in three days by 3 different users. We can observe that different persons have different environments. Probably, the most remarkable reason for being able to detect visually the differences in the users' datasets is usually due to the distribution and aspect of scenes, objects and people that appear. Following these premises, in this paper, we address the problem of automatically discovering which are the usual objects that form the environment of a person wearing the camera by means of a novel Object Discovery (OD) method. We must note the difference between \textit{Object Recognition}, where the goal is to discriminate objects according to their  classes by a classifier previously trained with a set of training samples; \textit{Object Detection}, where we should detect the subregion in the image where an object appears; and \textit{Object Discovery}, where we have to both detect new object instances or concepts, and assign them a label even without having training examples from all possible classes of objects.

\subsection{Previous Work}

Several works have been previously done in the OD field, some using segmentation techniques \citep{schulter2013unsupervised,russell2006using}, others extracting objects relying on visual words \citep{russell2006using,sivic2005discovering,liu2007unsupervised}. In \citep{chatzilari2011semi}, a semi-supervised method for segmentation-level labeling is presented and in \citep{tuytelaars2010unsupervised} a comparison of unsupervised OD methods is shown. One of the best performing OD methods is the one Lee's et.al.  published in \citep{lee2011learning}, where the authors propose a semi-supervised OD approach for object discovery. It starts by selecting the easiest objects by an objectness detector and keeps an iterative discovery procedure by clustering object candidates, selecting the best one as the one corresponding to the newly discovered object and applying an One-Class SVM to discover harder instances of it. The authors use a set of low-level image appearance (texture, colour and shape) and context features. One of its main drawbacks is that the features that it used are not rich enough to capture the characteristics of any existent real world object. More recently, in \citep{kading2015active}, a method for object discovery relying in active learning was presented. The authors base their work in the assumption that when dealing with an active learning problem, the oracle does not always know all the classes in advance and that, furthermore, not all the classes are always interesting for the problem at hand. With this in mind, they propose an Expected Model Output Change (EMOC) criterion for selecting the most relevant and useful images to label for the problem they are addressing, and at the same time trying to avoid no valid objects by using a local density measure. Cho et al. in \citep{cho2015unsupervised} worked on a part-based object discovery by proposing a new probabilistic matching strategy (Probabilistic Hough Matching) based on HOG descriptors for finding similar objects in different images. Additionally, they propose an associated confidence for finding the most outstanding object in each image.

In egocentric data, object discovery has been studied in much less extent. There, the OD  brings new challenges considering the non-intentionallity of the images, that is, compared to usual intentional images, the objects and people (if any) usually do not appear in centered positions, and partial occlusions produced by other objects or the image border are quite frequent. %The authors in \citep{herbst2011rgb} use an RGB-D camera and create a set of 3-D maps for distinguishing the different object classes.
% , even though, considering there is no wearable camera with these capabilities their method could not be applied in our field.
In \citep{kang2011discovering}, the authors define a method for finding new objects that a person can encounter in their daily living. They start by applying a segmentation of the images at different levels, extracting colour, texture and shape information from each segment and applying a series of grouping and refinement steps to find consistent clusters that can represent new concepts.
The authors in \citep{fathi2011learning} develop an object recognition method that uses segmentation techniques for extracting objects on egocentric visual data. In this case, the data acquired is captured using head-mounted cameras with high-temporal resolution (about 30 fps), what makes impossible to record the whole day of the person (due to memory and battery constraints). In order to solve this problem, we use cameras with low-temporal resolution (2-3 frames per minute) that are worn on chest level for maximizing the user comfort. As a result, we obtain a collection of images instead of a video, where objects are captured non-intentionally, and frequently appear blurred and non-centred. The main additional challenges these cameras cause are: 1) having frames so much temporally spaced disable the possibility to directly infer information from sequential frames and 2) extracted motion information is not reliable enough.

The main handicaps of existent OD methods are: 1) they lack a way to capture and reuse the knowledge acquired when analyzing the previous data, which is very important considering the redundancy  of the data acquired in lifelogging \citep{min2014efficient}, and 2) many OD methods rely on using as a first step an object detection algorithm like \citep{alexe2010object,cheng2014bing,arbelaez2014multiscale,uijlings2013selective} for having an initial set of object candidates. As we prove in section \ref{subsec:object_detection}, these methods usually produce a very high number of False Positives (FP) that should be dealt with.

\subsection{Contributions}

In this paper, we propose a new OD method for egocentric data (based on our previous work presented in \citep{bolanos2015object}), that we call Ego-Object Discovery (EOD). Our contributions start by using a set of powerful features extracted by means of a Convolutional Neural Network (CNN). These networks are proving their huge potential to address different problems in the field of Computer Vision (\citep{Lee2009convolutional,Lee2009Unsupervised,Goodfellow2014Multi}, just to mention a few). Lately, a new method \citep{moghimiexperiments} using CNN data has been proposed for egocentric activity recognition. However, no methods on OD using these features exist yet. To overcome the problem present in previous works of nonexistent knowledge reuse we use a new Refill methodology, which allows to discover new samples from the categories, even having a low number of instances, which are quite present in egocentric sequences. As additional contributions w.r.t. our previous work, we here present a strategy for solving the high number of FPs (or 'No Object' candidates) produced by the object detection methods: a SVM filtering strategy. Also introduce the first egocentric object discovery dataset (EDUB) of lifelogging data with ground truth (GT) object segmentations, apply a comparison with the state of the art object detection algorithms, and analyze the results of our method also on two public datasets of intentional images (PASCAL and MSRC).%  (Fig. \ref{fig:datasetsamples}).

The article is organized as follows: in section \ref{sec:approach}, we define the EOD  algorithm. In section \ref{sec:results}, we present the datasets used to validate our method, the tests of EOD on all datasets, comparison of state of the art object detectors and discussions on the obtained results. We finish with some conclusions and future work.

%\begin{figure}[ht]
%  \centering
%  \includegraphics[width=1\columnwidth]{datasetsamples2.png}
%  \caption{Images acquired by a lifelogging device, where objects of interest appear like: person, TV monitor, paper, bycicle, face, sign, building, hand, chair, etc.}
%  \label{fig:datasetsamples}
%\end{figure}

%========================================================%
%========================================================%
%				APPROACH
%========================================================%
%========================================================%
\section{The Ego-Object Discovery Approach} \label{sec:approach}

\begin{figure}[!ht]
  \centering
  \includegraphics[width=\columnwidth]{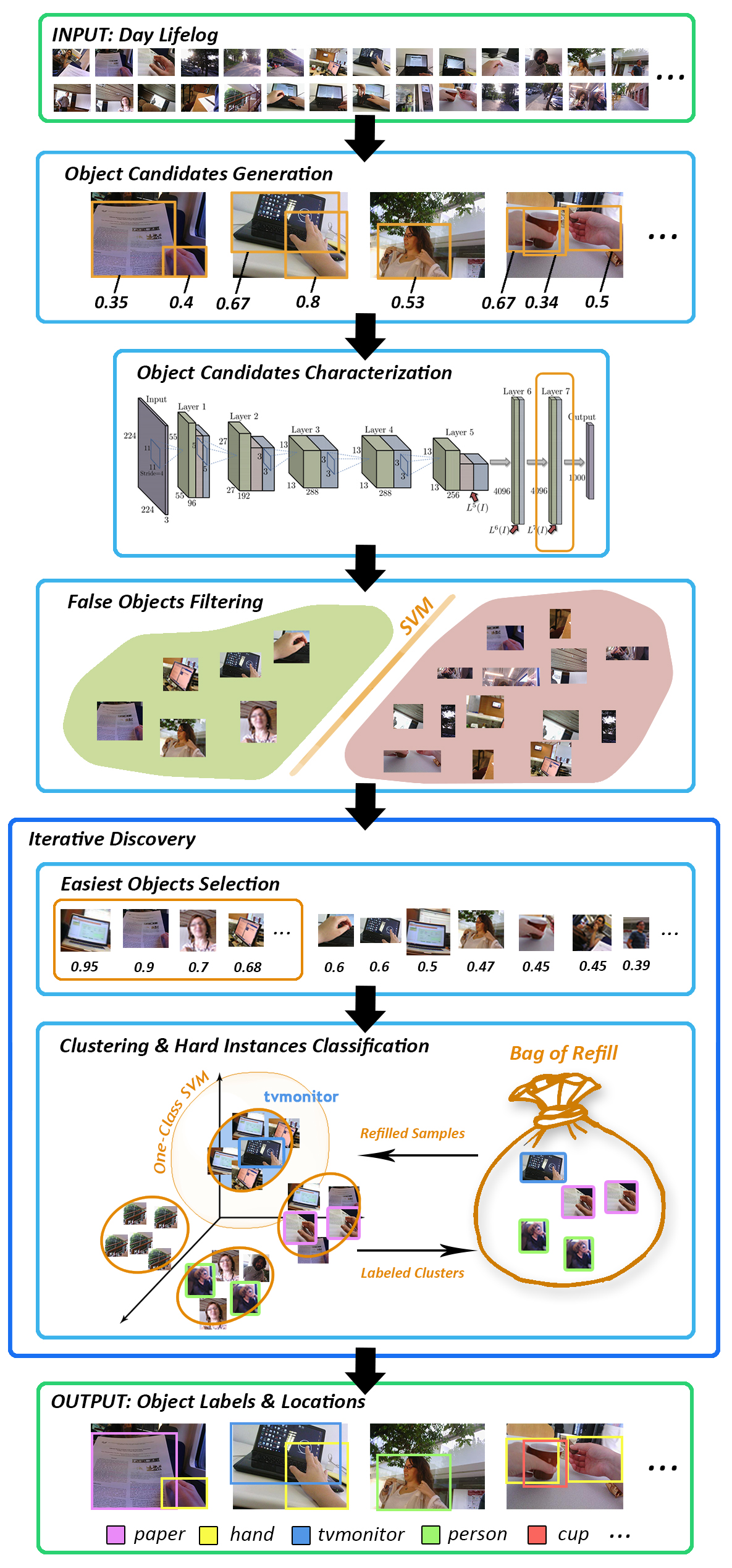}
  \caption{Ego-Object Discovery methodology scheme. The different algorithms applied in each part of the methodology are represented in orange.}
  \label{fig:algorithm}
\end{figure}

Given the problem of OD in low-temporal resolution egocentric data, our algorithm is formulated as an iterative procedure. At the beginning, it should be provided with a seed of initial objects information to expand, defined as a small bag of labeled objects, represented by their regions, and called a {\bf bag of refill}. The EOD algorithm passes through several steps (see Fig. \ref{fig:algorithm}): a) it detects image regions representing object candidates and their corresponding objectness scores from each new set of images, b) extracts object candidates features by using a pre-trained CNN, c) filters false object ('No Object') instances and d) proceeds with a clustering-based iterative procedure as follows: 1) on the {\em easiest} objects, it applies a {\em refill strategy} by using the bag of refill, 2) clusters them by using an agglomerative clustering approach and labels the best cluster that represents the newly discovered object and 3) applies a supervised expansion to find harder instances of it. After a fixed number of $t$ iterations or until no easy sample remains, it outputs the set of found object coordinates and labels. 

To describe and cluster the candidates, EOD uses both appearance and local context features. Appearance are extracted by a CNN  \citep{Jia13caffe}, and context is provided by both the inherent description of the object background that also extracts the CNN, and indirectly the refill procedure, that will introduce instances of the same classes but with different backgrounds. Being very suitable for lifelogging images considering the redundancy of the objects we routinely see. In the following subsections, we give details about each step of the EOD procedure.

%\noindent
%{\it Algorithm for Object Discovery}
%\begin{verbatim}
%INPUT: photos/video sequence.
%OUTPUT: object labels and locations in the input video.
%PROCEDURE:
%    candidates := extractObjectnessScore(video)
%    do {
%        easyObjects := getEasiestObjects(candidates)
%        easyObjects := doRefill(candidates, easyObjects)
%        appearanceFeatures := getCNNAppearance(easyObjects)
%        clusters := clusterizeWard(appearanceFeatures)
%        bestCluster := getBestSilhouetteCoeff(clusters)
%        label(bestCluster)
%        model := buildOneClassSVM(appearanceFeatures, bestCluster)
%        classifyHarderInstances(appearanceFeatures, model)
%    while(easyObjects)}
%\end{verbatim}
%\vspace{-1em}

\subsection{Object Candidates Preparation}

%%%%%%%%%%%%%%%%%%%%%%%%%%%%%%%%%%%%%%%%%%%%%%%%%%%%%%
%			Objectness and Easiness
%%%%%%%%%%%%%%%%%%%%%%%%%%%%%%%%%%%%%%%%%%%%%%%%%%%%%%
%\subsection{Object Candidates found by Objectness Detector} \label{sec:easiness}
%\subsection{Object Candidates Generation} \label{subsec:objvsnoobj_svm}

\vspace{1em}

{\bf Object Candidates Generation:} The first step needed to characterize the environment of the user through object discovery is extracting a set of object candidates for each image. To do so, we used the Objectness detector provided by Ferrari et al. in \citep{alexe2010object}, which additionally to the bounding box for each candidate, outputs a score associated to the probability of being a true object (objectness score). This score is produced by three visual cues: \textit{Multi-scale Saliency} (finds blob-like structures at multiple scales that could indicate the presence of an object); \textit{Color Contrast} (finds high colour differences between the analyzed bounding box and its surroundings); and \textit{Superpixels Straddling} (penalizes the bounding boxes that do not respect the boundaries of the superpixels in the image).

\vspace{1em}
%%%%%%%%%%%%%%%%%%%%%%%%%%%%%%%%%%%%%%%%%%%%%%%%%%%%%%
%			Features Used
%%%%%%%%%%%%%%%%%%%%%%%%%%%%%%%%%%%%%%%%%%%%%%%%%%%%%%
%\subsection{Features for Object Discovery} \label{subsec:features used}

{\bf Object Candidates Characterization:} As features to cluster the object candidates, we used a pre-trained CNN \citep{krizhevsky2012imagenet}, which was trained on millions of images and is composed as  a succession of convolutional and pooling layers. We deleted the last layer, which offers a supervised classification of    1.000 ImageNet classes, and used the output of the penultimate layer as our features (4096 variables). Note that our approach is different to the one of \citep{lee2011learning} that used: LAB histograms for extracting colour information, Pyramid HOG for extracting shape information, and Spatial Pyramid Matching \citep{lazebnik2006beyond} for extracting texture information.

\vspace{1em}

%%%%%%%%%%%%%%%%%%%%%%%%%%%%%%%%%%%%%%%%%%%%%%%%%%%%%%
%			ObjVSNoObj SVM Filter
%%%%%%%%%%%%%%%%%%%%%%%%%%%%%%%%%%%%%%%%%%%%%%%%%%%%%%
% \subsection{SVM Filter} \label{subsec:objvsnoobj_svm}

{\bf False Objects Filtering:} The main drawback of most object detection methods is the huge number of FPs, they produce.
% Added to the fact that we have a limited amount of time to label as many true object instances as possible, 
Given that it is not enough to rely on the objectness score for discarding the 'No Object' instances, we filter the object candidates by an RBF-SVM classifier trained  on CNN features to distinguish 'Object' vs. 'No Object' instances.

\subsection{Iterative Discovery}

\vspace{1em}

{\bf Easiest Objects Selection:} In order to achieve an iterative easy-first discovery, we used their associated objectness score to decide if a candidate $\omega$ is considered in the current iteration: 

\begin{equation}
    objectnessScore(\omega) > \mu + \omega_1 \sigma - \omega_2 t ,
\end{equation}    

where $\mu$ and $\sigma$ are respectively, the mean and the standard deviation of all scores, $t$ is the current iteration, and $\omega_1$ and $\omega_2$ are weights. This easiness measure seems a promising method for obtaining object candidates in general. However, this technique does not obtain the same results in egocentric datasets than in intentional images due to the fact that images are not captured by a person looking at objects of the world, but are acquired non-intentionally while a person is loosely wearing the camera. As a result of the inherent low frequency of appearance of different objects of the real world, to the limited image quality of the wearable egocentric devices and to the constant moving of the user, a great part of the photos are unclear, dark or blurry (see Fig. \ref{fig:objects_person}). All this causes lower precision, when clustering the obtained object candidates. 

\vspace{1em}

%%%%%%%%%%%%%%%%%%%%%%%%%%%%%%%%%%%%%%%%%%%%%%%%%%%%%%
%			Refill Strategy
%%%%%%%%%%%%%%%%%%%%%%%%%%%%%%%%%%%%%%%%%%%%%%%%%%%%%%
% \subsection{Object Discovery} \label{subsec:refill}

\begin{figure}[h]
\centering
\begin{minipage}[b]{0.45\linewidth}

	\center
	\includegraphics[width=\columnwidth]{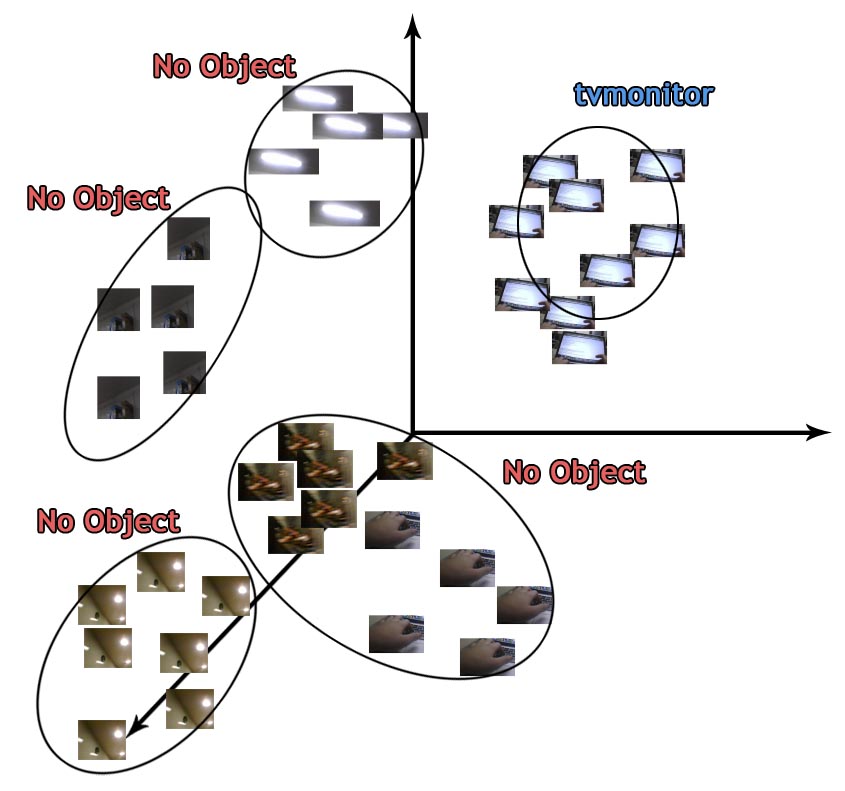} 
    \caption{Clusters formed by the easiest samples.}
    \label{fig:refill1}

\end{minipage}
\quad
\begin{minipage}[b]{0.45\linewidth}

	\center
	\includegraphics[width=\columnwidth]{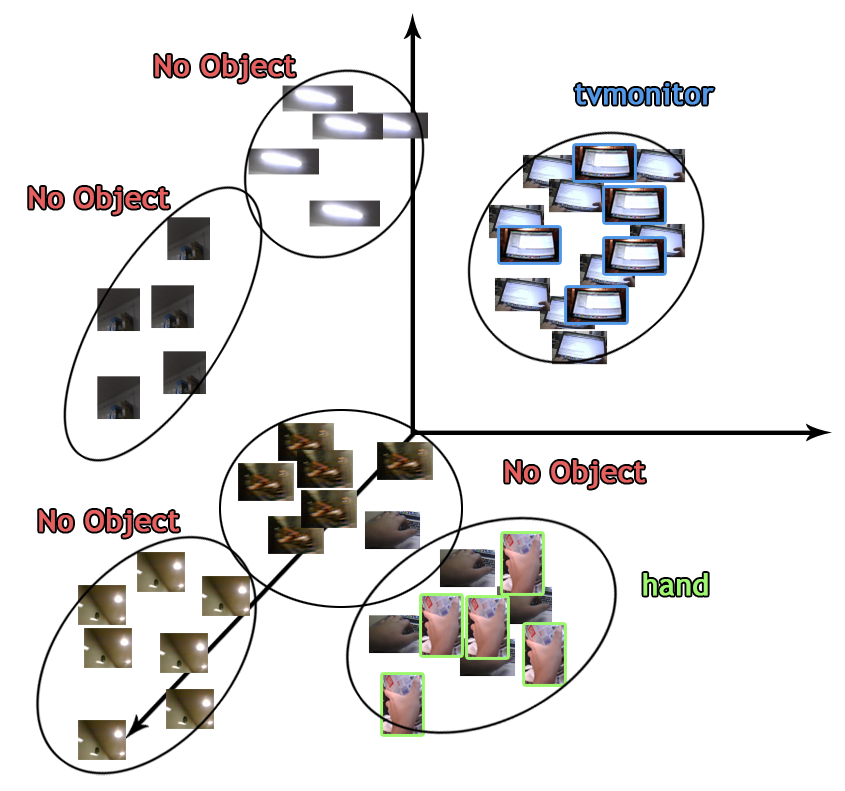} 
    \caption{Clusters formed by the refilled and easiest samples.}
    \label{fig:refill2}

\end{minipage}
\end{figure}

{\bf Refill Strategy:} In order to solve these problems, we define a "refill" methodology as follows: at each iteration, the set of selected easiest samples is completed with a certain percentage (w.r.t. the number of easy samples retrieved) of samples from the Bag of Refill, which are randomly chosen labeled samples distributed on the already discovered object classes. In this way, we address two problems: 1) difficulty to form a cluster from a very small set of class instances, and 2) difficulty to link samples of the same class that were blurry and unclear. So, refilling the space with more samples of the same class of objects, we can obtain more compact clusters (see Fig. \ref{fig:refill1} and Fig. \ref{fig:refill2}).

%%%%%%%%%%%%%%%%%%%%%%%%%%%%%%%%%%%%%%%%%%%%%%%%%%%%%%
%	  Clustering and Hard Instancess Classification
%%%%%%%%%%%%%%%%%%%%%%%%%%%%%%%%%%%%%%%%%%%%%%%%%%%%%%
%\subsection{Object Discovery} % Clustering and Hard Instancess Classification}

{\bf Clustering and Hard Instances Classification:} In this step we apply an Agglomerative Ward clustering on the object candidates. %, where we used as cutoff clustering criterion $\text{cutoff} = 2 \sigma^2 + \mu$, being $\sigma^2$ and $\mu$ the standard deviation and the mean of all the distances between the clusters in the resulting hierarchy.
Moreover, once the clusters are formed, we get the Silhouette Coefficient \citep{tan2011v} on each cluster and select the best for the user to assign it a label. This coefficient is only calculated on the unlabeled samples, never using the refilled ones for selecting the most reliable cluster. At the end of each iteration, a OneClass-SVM %(with $\nu=0.1$) 
for searching for harder instances is built with the new cluster  and the rest of the easy samples are classified. %In any case, the refilled samples, which were already labeled, can only change their labels if did not belong to the initial bag of refill.

%========================================================%
%========================================================%
%				RESULTS
%========================================================%
%========================================================%
\section{Results} \label{sec:results}

In this section, we discuss the three datasets we used (summarizing their characteristics in Table \ref{tab:num_samples}), and expose the different tests applied to illustrate the EOD performance.

%%%%%%%%%%%%%%%%%%%%%%%%%%%%%%%%%%%%%%%%%%%%%%%%%%%%%%
%	  Datasets
%%%%%%%%%%%%%%%%%%%%%%%%%%%%%%%%%%%%%%%%%%%%%%%%%%%%%%
\subsection{Datasets} \label{sec:datasets}

Due to the low number of publicly available egocentric datasets and the complete lack of egocentric object-labeled datasets, we considered very important to construct one and make it public in order to serve as a base for algorithms comparison for the egocentric community.

\vspace{1em}

\begin{figure}[h]
  \centering
  \includegraphics[width=\columnwidth]{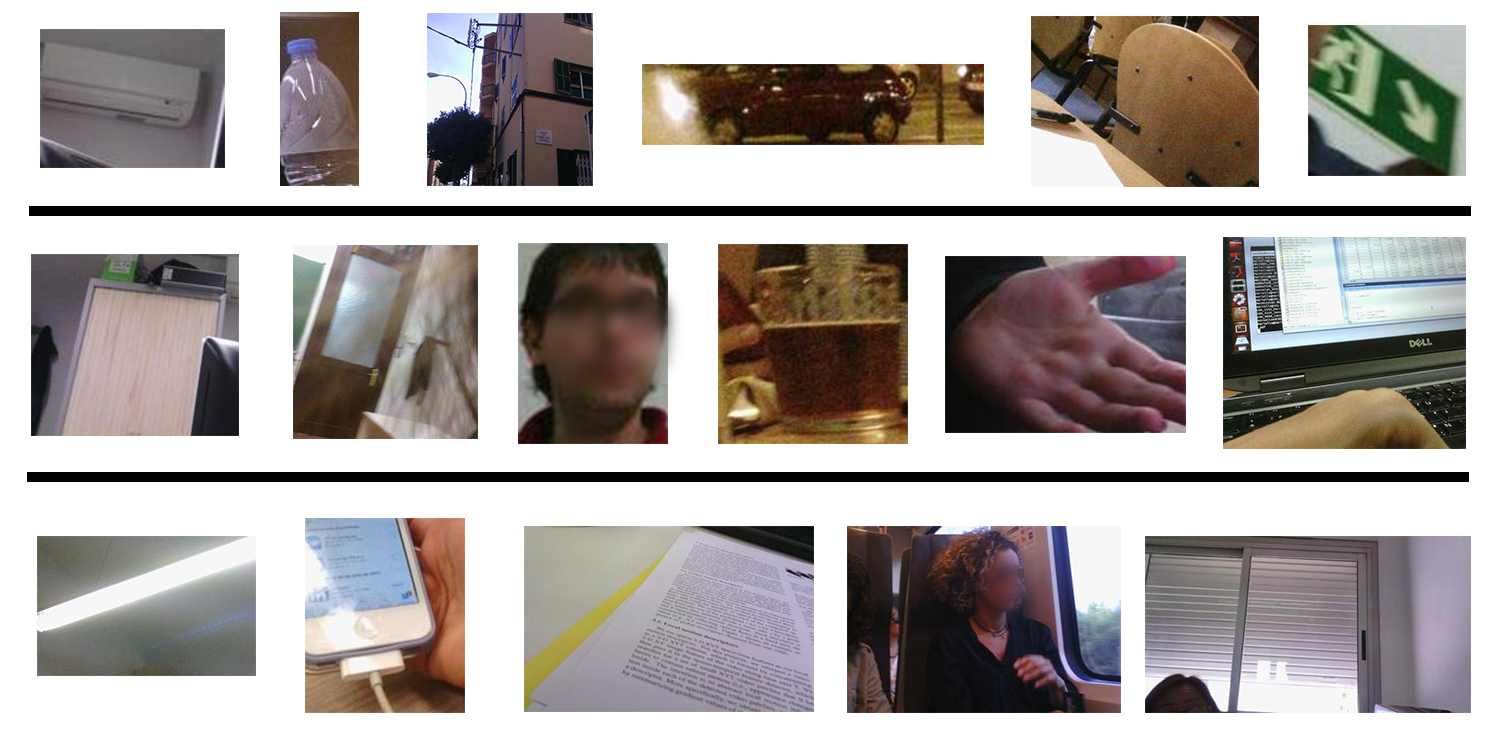}
  \caption{Object candidates obtained by the Ferrari's objectness detector on the EDUB dataset. From left to right and top to bottom: aircon, bottle, building, car, chair, sign, cupboard, door, face, glass, hand, tvmonitor, lamp, mobilephone, paper, person, window.}
  \label{fig:EDUB_samples}
\end{figure}

The \textbf{Egocentric Dataset of the University of Barcelona (EDUB)} (see Fig. \ref{fig:EDUB_samples}) is a dataset composed of 4912 images acquired by 4 people using the Narrative wearable camera (www.getnarrative.com). It is divided in 8 different days, 2 days per person. The objects appearing in the images were segmented using the online tool LabelMe \citep{russell2008labelme} (although here we only use their bounding box) and their annotation files are similar to the ones provided by PASCAL. EDUB includes the following classes (number of samples per class are given in parenthesis): 'lamp' (2299), 'tvmonitor' (1274), 'hand' (1232), 'person' (1175), 'glass' (831), 'building' (732), 'face' (565), 'aircon' (530), 'sign' (506), 'cupboard' (392), 'paper' (377), 'car' (315), 'bottle' (260), 'door' (199), 'chair' (179), 'mobilephone' (145), 'window' (138), 'dish' (65), 'motorbike' (64), 'bicycle' (12), and 'train' (4). Note that in our tests, we did not use the classes with few instances (i.e. smaller than 100), considering that it would not be possible to discover them with a clustering strategy.

The second of the datasets we considered is the \textbf{PASCAL VOC 2012} \citep{pascal-voc-2012}, being one of the most widely used in object detection/recognition research , with very difficult and challenging images. We used the 'trainval' (for having more samples) set of images for our tests, but previously deleted the images that had in common with its 2007 version. We applied this pre-processing to avoid any bias in the results, since some of the used object detection methods were trained using PASCAL VOC 2007.

\vspace{1em}

The last of the datasets, we chose is the \textbf{Microsoft Research Cambridge (MSRC)} \citep{msrc}, which was also used in \citep{lee2011learning} for object discovery, and therefore will ease the comparison of the results. 
Considering that MSRC dataset is labeled at pixel level, we had to extract the bounding boxes corresponding to each of the objects making some assumptions: 1) the bounding box for an object is the minimal closing box around all the connected pixels that belong to the same class; 2) given the dataset is split in folders, we only considered valid the objects with the same class as the folder's name; 3) the minimal area for an object to be valid was set to 50x50 image pixels (about 0.81\% of the whole image); and 4) we excluded the labels 'grass', 'sky', 'mountain', 'water' and 'road', because they are not objects, but rather environments.

\begin{table}[]
\centering
\caption{Image/object characteristics for each of the used datasets.}
\vspace{0.5em}
\begin{tabular}{|c||c|c|c|c|} \hline
 & \textit{images} & \begin{tabular}{@{}c@{}} \textit{object} \\ \textit{candidates}\end{tabular} & \textit{GT objects} & \textit{classes} \\ \hline \hline
 \textit{MSRC} & 3,427 & 171,350  & 4,217 & 16 \\ \hline
 \textit{PASCAL} & 16,369 & 818,450 & 38,144 & 20 \\ \hline
 \textit{EDUB} & 4,912 & 245,600 & 11,149 & 17 \\ \hline
\end{tabular}
\label{tab:num_samples}
\end{table}

\begin{figure}[ht]
  \centering
  \includegraphics[width=\columnwidth]{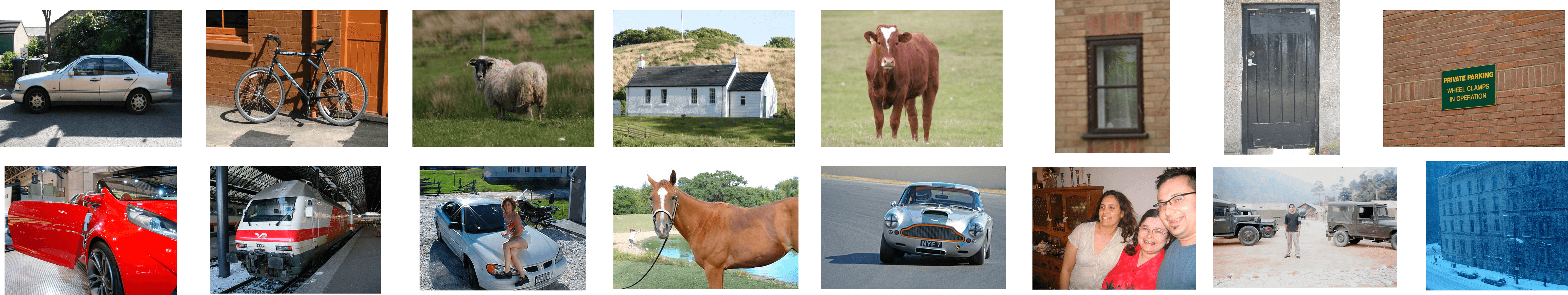}
  \caption{MSRC image samples (top) and PASCAL 12 samples (bottom).}
  \label{fig:MSRC_PASCAL_samples}
\end{figure}

Fig. \ref{fig:objects_person} and \ref{fig:MSRC_PASCAL_samples} show some image samples from the 3 datasets. MSRC dataset, compared to the other two should obtain better results due to the position of the objects (central to the image) and their clear appearance. Even though in general PASCAL has some object instances very difficult to find, the hardest one is the EDUB (also considering the high rate of objects occlusions, blurriness and lower image quality).

%%%%%%%%%%%%%%%%%%%%%%%%%%%%%%%%%%%%%%%%%%%%%%%%%%%%%%
%	  Object Detection Methods
%%%%%%%%%%%%%%%%%%%%%%%%%%%%%%%%%%%%%%%%%%%%%%%%%%%%%%
\subsection{Object Detection Methods}\label{subsec:object_detection}

\vspace{1em}

Given that the first step of the algorithm is to obtain object candidates from the images, we tested and compared four different state of the art object detection methods on the three datasets (see Table \ref{tab:object_detection_methods}). We chose Objectness \citep{alexe2010object}, BING \citep{cheng2014bing}, Multiscale Combinatorial Grouping (MCG) \citep{arbelaez2014multiscale} and Selective Search \citep{uijlings2013selective} methods considering their good performances. For MCG, we applied its quickest, but less exhaustive version.

Due to the dramatic increase of space needed to store all the samples\footnote{Considering the PASCAL 12 dataset, we needed nearly 30GB of data to store all the images and features for the tests}, we extracted the top $W=50$ object candidates per image sorted by their objectness score. 
% Although, we must also consider that increasing the value of $W$ would improve the Detection Rate (DR) and consequently the F-Measure results of our framework.

\begin{table}[]
\centering
\caption{Percentage of 'No Objects' (NO) (or of False Positives) and Detection Rate (DR) comparison of the four object detection methods on our three datasets.}
\vspace{0.5em}
\begin{tabular}{|c||c|c|c|c|c|} \hline
 &  & \textit{Objectness} & \textit{BING} & \textit{MCG} & \textit{Sel.Search} \\ \hline \hline
 \textit{MSRC} & \begin{tabular}{@{}c@{}}\footnotesize{NO} \\ \footnotesize{DR}\end{tabular} & \begin{tabular}{@{}c@{}}91.69 \\ \textbf{88.83}\end{tabular} & \begin{tabular}{@{}c@{}}96.68 \\ 64.15\end{tabular} & \begin{tabular}{@{}c@{}}\textbf{48.42} \\ 79.61\end{tabular} & \begin{tabular}{@{}c@{}}61.95 \\ 70.98\end{tabular} \\ \hline
 \textit{PASCAL} & \begin{tabular}{@{}c@{}}\footnotesize{NO} \\ \footnotesize{DR}\end{tabular} & \begin{tabular}{@{}c@{}}92.14 \\ \textbf{60.47}\end{tabular} & \begin{tabular}{@{}c@{}}92.93 \\ 56.93\end{tabular} & \begin{tabular}{@{}c@{}}\textbf{65.16} \\ 49.36\end{tabular} & \begin{tabular}{@{}c@{}}71.30 \\ 36.71\end{tabular} \\ \hline
 \textit{EDUB} & \begin{tabular}{@{}c@{}}\footnotesize{NO} \\ \footnotesize{DR}\end{tabular} & \begin{tabular}{@{}c@{}}92.75 \\ \textbf{60.45}\end{tabular} & \begin{tabular}{@{}c@{}}95.43 \\ 50.00\end{tabular} & \begin{tabular}{@{}c@{}}\textbf{79.17} \\ 49.57\end{tabular} & \begin{tabular}{@{}c@{}}84.27 \\ 29.09\end{tabular} \\ \hline
\end{tabular}
\label{tab:object_detection_methods}
\end{table}

Analyzing the percentage of NO (see overlapping score in section \ref{subsec:test_settings}) and DR of each method, we can see that the DR is not as high as desired and the \% of NO is remarkably high. Meaning that using any of the best state of the art approaches for object detection makes us lose a lot of information, so we have to consider that our final results will be inevitably biased and worsen for this reason.

Comparing the different datasets, as one could immediately expect looking at the images, it is clearly easier for any objectness measure to get good results on the MSRC dataset, meanwhile it is quite more difficult on PASCAL and EDUB, having an extra difficulty for the second one due to the non-intentional acquisition and less clear images of the wearable cameras.

Given our final goal of being able to discover the true distribution of object classes and as many individual GT objects as possible, we considered that the objectness measure that obtained better results for EOD  was the one proposed by \citep{alexe2010object}, 
% \footnote{In the code we made available, there is also the possibility to extract object candidates and testing the OD process with any of the four  object detection methods},
because we are interested in getting most of the GT objects in the dataset, even if we have to deal with a lot of NO (i.e. noisy or FP) instances.

%%%%%%%%%%%%%%%%%%%%%%%%%%%%%%%%%%%%%%%%%%%%%%%%%%%%%%
%	  Experimental Setup
%%%%%%%%%%%%%%%%%%%%%%%%%%%%%%%%%%%%%%%%%%%%%%%%%%%%%%
\subsection{Experimental Setup}\label{subsec:test_settings}

\vspace{1em}

In order to perform the methodology validation, we first \textbf{leave a 50\% of the object classes in the unlabeled pool} as a test set. Note that we need to test if the algorithm is able to discover unseen object classes. From the remaining part of classes, similar to \citep{lee2011learning}, we separated a 40\% of the total object candidates to represent the initial knowledge located into the bag of refill and used the remaining 60\% for testing, too.

In order to say that a candidate matches a GT object bounding box, we followed the PASCAL VOC challenge criterion, that uses the Overlapping Score (OS). Given a window region $\omega$ produced by the object detector, is considered a hit on a GT label, iff:
% the OS between its bounding box and the detected object region: 
\begin{equation}
	OS = \frac{|GT \cap \omega|}{|GT \cup \omega|}>0.5
\end{equation}
Due to the challenging images presented to the object detector, a very high percentage of samples (more than 92\% using Ferrari's objectness) could not be considered objects, and were labeled as NO.

In order to tune the parameters for the SVM filter strategy for each of the datasets, we applied a nested 5-fold cross-validation with 5 test divisions with a grid of parameters of $\sigma \in \{0.1, 0.5, 3, 10, 100, 1000\}$ and $C \in \{0.1, 0.5, 3, 10, 100, 1000\}$. All the tests were performed for each dataset separately and on a randomly selected fraction of its samples to save computational time. With these tests, we finally found that the best parameters for filtering as many NO instances and at the same time keeping as many 'Object' instances as possible (high sensitivity and high specificity) for both the PASCAL and the MSRC classifiers were $\sigma = 100$ and $C = 3$. In the labeling step, for simulation purposes, we labeled the best cluster with a majority voting strategy w.r.t the GT, although this labeling is intended to be made by the camera user his-/herself.

We designed different test settings to evaluate our proposal:
\begin{description}
    \item{S1:} Features of \citep{lee2011learning}.
    \item{S2:} CNN object features.
    \item{S3:} CNN object features with Refill  strategy.
    \item{S4:} CNN object concatenated with CNN scene features and Refill strategy.
    \item{S5:} CNN object features with Refill and SVM filter.
    \item{S6:} CNN object features with Refill, SVM filter and PCA.
    % 99\% variance.
\end{description}

With the first pair of settings, we intend to compare the generalization capabilities of the appearance features from \citep{lee2011learning} against the extracted CNN features. In setting S4, we tested adding a context about the scene, and in setting S6, we applied a PCA feature dimensionality reduction and transformation in case there is redundancy in the extracted CNN features.

%%%%%%%%%%%%%%%%%%%%%%%%%%%%%%%%%%%%%%%%%%%%%%%%%%%%%%
%	  Tests Comparison
%%%%%%%%%%%%%%%%%%%%%%%%%%%%%%%%%%%%%%%%%%%%%%%%%%%%%%
\subsection{Silhouette Coefficient Comparison}

\vspace{1em}

\begin{figure}[ht]
    \centering
	\includegraphics[width=\columnwidth]{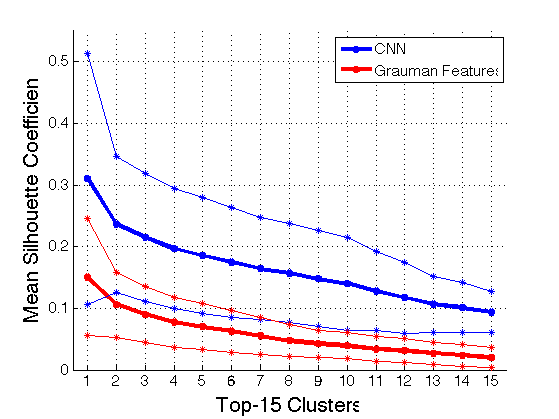} 
    \caption{Comparison of mean silhouette coefficient (thick lines) and standard deviation (thin lines) for the top 15 clusters on 50 algorithm iterations (high values are better). 
    }
    \label{fig:silhouette}
\end{figure}

In order to check if the clusters formed by using CNN features are more robust than the ones formed by using the features from \citep{lee2011learning}, we can analyse the mean silhouette coefficient values obtained in several iterations. In Fig. \ref{fig:silhouette} we plot the difference on the silhouette coefficient values obtained by using the two kind of features. The comparison is applied for the top 15 clusters on the first 50 iterations of the algorithm. 

We can immediately realise that the average compactness of the clusters and their difference to the other clusters (which is what Silhouette Coefficient measures) is always higher when using CNN features, and will lead to get purer clusters and a better labeling.

\subsection{F-Measure Comparison on EDUB} \label{subsec:edub_comparison}

\vspace{1em}

To evaluate our approach, we used the F-Measure, because it objectively penalizes the FP and FN objects in each class, that is, represents a trade-off between the Precision and Recall of the method. At the same time, we want to give the same importance to all classes, and  are interested in finding as many different classes as possible, but always leaving the NO instances aside, without considering them into the quality measures. Hence, we applied the average per-class precision and recall defined in \citep{sokolova2009systematic} in order to obtain the average F-Measure:

\begin{equation}
	\text{F-Measure} = 2 \frac{Precision_M * Recall_M}{Precision_M + Recall_M},
\end{equation}
where $Precision_M$ and $Recall_M$ are the mean precision and recall of all classes, giving the same weight to all of them.

\begin{figure}[h]
\centering
\begin{minipage}[b]{0.45\columnwidth}

	\center
	\includegraphics[width=\columnwidth]{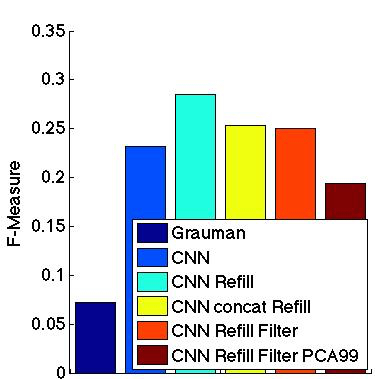} 
    \caption{Final F-Measure for each setting.}
    \label{fig:measures_final}

\end{minipage}
\quad
\begin{minipage}[b]{0.45\columnwidth}

	\center
	\includegraphics[width=\columnwidth]{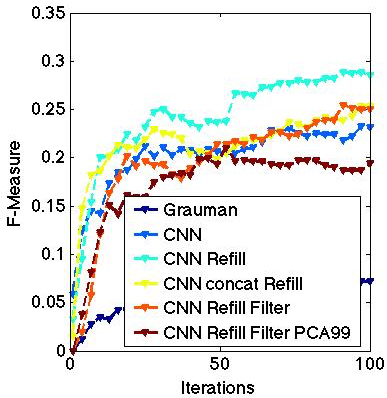} 
    \caption{F-Measure evolution for each different setting.}
    \label{fig:f_measure_evolution}

\end{minipage}
\end{figure}

All measures were averaged by at least 5 executions per setting and for a maximum of 100 algorithm iterations. Using these tests, we compared all settings at the end of the easiest samples discovery (Fig.\ref{fig:measures_final}) and on each iteration (Fig.\ref{fig:f_measure_evolution}).

%\begin{figure}[ht]
%    \centering
%	\includegraphics[width=0.8\columnwidth]{measures_EDUB} 
%    \caption{Final F-Measure, Purity and Accuracy for each setting}
%    \label{fig:measures_final}
%\end{figure}

%\begin{figure}[ht]
%	\centering
%	\includegraphics[width=0.8\columnwidth]{f_measure_EDUB} 
%    \caption{F-Measure evolution for each different setting}
%    \label{fig:f_measure_evolution}
%\end{figure}

Looking at Fig.\ref{fig:measures_final}, we can clearly see that using CNN outperforms the features of \citep{lee2011learning}, indicating that they can form purer clusters and find a wider variety of classes thanks to their best representation. Then, adding the Refill technique, the EOD method outperforms the one using the CNN features only.
%, which proves the assumptions made in section \ref{subsec:objvsnoobj_svm}. 
The rest of the methods can not reach the same results as CNN + Refill. Moreover, using the additional CNN features of the whole image adds just noise to the set of features. That is, simply by using the CNN with the bounding box of the object candidate already captures the closest and most relevant object context. Considering the high dimensionality of CNN features, it seems that including a PCA dimensionality reduction to the data does not provide any benefit to the object discovery.

Comparing the evolution of the F-Measure through the iterations (Fig. \ref{fig:f_measure_evolution}), we see that any of the settings using CNN features experiments a much higher increase in the F-Measure value just in the first 5-10 iterations, meaning that they can find clusters of true objects quicker than using the setting S1.

Also, using the CNN features combined with the refill strategy, the results clearly improved from 0.072 to 0.285. This is caused by the discovery of different classes of samples. While when using the features of \citep{lee2011learning}, we are only able to discover 3 or 4 classes at most, achieving an average of 0.072 F-Measure; with the setting S3, we can discover instances of more than half of the classes, getting nearly 0.29 of F-Measure. Although on the EDUB using the setting S5 (CNN + Refill + SVM Filtering) does not seem to get as good F-Measure results as on the other settings, in other datasets, as we will be able to see, it outperforms or nearly reaches the results of setting S3. Furthermore, it gets a wider variety of object classes.

\subsection{F-Measure Comparison on All Datasets} \label{sec:f_measure_all}

\vspace{1em}

After having found the best combination of methods and parameters to use, we tested and compared how good the new method was contrasting it with  the state of the art method \citep{lee2011learning}  for any of the datasets (EDUB, PASCAL 2012 and MSRC). In table \ref{tab:summary_results_comparison}, we can see a summary of the F-Measure results obtained for each of the datasets and each of the best test settings (average on at least 5 tests per setting).

\begin{table}[ht]
\centering
\caption{F-Measure comparison for the three datasets, the state of the art \citep{lee2011learning} and our best test settings (CNN + Refill and CNN + Refill + Filter).}
\vspace{0.5em}
\begin{tabular}{|c||c|c|c|} \hline
F-Measure & S1 & S3 (ours) & S5 (ours) \\ \hline \hline
 \textit{MSRC} & 0.121 & \textbf{0.431} & 0.410 \\ \hline
 \textit{PASCAL} & 0.002 & 0.145 & \textbf{0.179} \\ \hline
 \textit{EDUB} & 0.072 & \textbf{0.285} & 0.250 \\ \hline \hline
 \textit{Average} & 0.065 & 0.287 & 0.280 \\ \hline
\end{tabular}
\label{tab:summary_results_comparison}
\end{table}

As we can see, using any of our best methods (either setting S3 or setting S5) clearly outperforms the state of the art features, having from a 350\% to a 9000\% of improvement depending on the dataset and the settings, and a 453\% of average improvement with the best setting.

Even though the average F-Measure result obtained using the SVM filtering (setting S5) is worse than without it (setting S3), we must consider that these classifiers have been built with samples from different datasets than the ones on test (1/2 of the PASCAL samples for MSRC tests and all MSRC samples for both PASCAL and EDUB tests), meaning that the generalization will be poorer than if we built a general classifier with images from any of the datasets.

Another important consideration we must take into account, is that for the MSRC tests, although the final (after 100 iterations) F-Measure results are better without the filtering, in fact they were better with the filtering from the 1st to the 75th iteration, meaning that in some cases, it can offer better results if we want to stop early the discovery method.

\subsection{Object Discovery Results} \label{sec:object_discovery_figures}

\vspace{1em}

\begin{figure}[ht]
    \centering
	\includegraphics[width=\columnwidth]{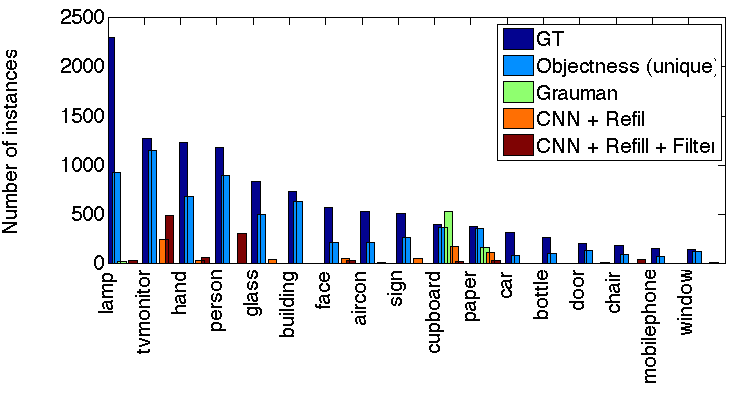} 
    \caption{Objects found by each method compared to the GT and the ones found by the Objectness measure \citep{alexe2010object}. 
    }
    \label{fig:objects_distribution}
\end{figure}

In this section, we analyze the object discovery results in more general terms. 
% In all the plots of this section, we have to consider that we are only using the results obtained by a single execution (randomly selected) for each of the methods.
In Fig. \ref{fig:objects_distribution}, we can see the absolute number of object instances found by each of the methods compared to the GT and the ones found by the Objectness measure (\citep{alexe2010object}, in this case without counting repeated instances of the same object).

As we can see, using the parameters of setting S1 \citep{lee2011learning}, we are only able to find instances from 3 different classes, which causes the previously seen very low F-Measure results. On the other hand, using either CNN + Refill (setting S3) or CNN + Refill + Filter (setting S5), we can clearly discover objects from a wider variety of classes, which also causes the higher resulting F-Measure. Moreover, we get a wider variety of classes with setting S5 (10 different classes) than with setting S3 (8 different classes).

\begin{table*}[ht]
\centering
\caption{Number of clusters found for each class using any of the settings S1, S3 or S5.}
\vspace{0.5em}
\scriptsize
\begin{tabular}{|c||c|c|c|c|c|c|c|c|c|c|c|c|c|c|c|c|} \hline
Test & No Object & hand & lamp & cupboard & car & glass & chair & face & door & window & tvmonitor & building & paper & person & mobilephone & sign \\ \hline \hline
S1 & 96 & 0 & 1 & 2 & 0 & 0 & 0 & 0 & 0 & 0 & 0 & 0 & 1 & 0 & 0 & 0 \\ \hline
S3 & 71 & 1 & 0 & 3 & 0 & 1 & 0 & 6 & 0 & 0 & 8 & 0 & 4 & 0 & 0 & 3 \\ \hline
S5 & 49 & 2 & 3 & 6 & 0 & 0 & 4 & 5 & 1 & 1 & 23 & 0 & 1 & 5 & 0 & 0 \\ \hline
\end{tabular}
\label{tab:num_clusters_classes}
\end{table*}

\begin{figure*}[ht]
    \centering
	\includegraphics[width=\textwidth]{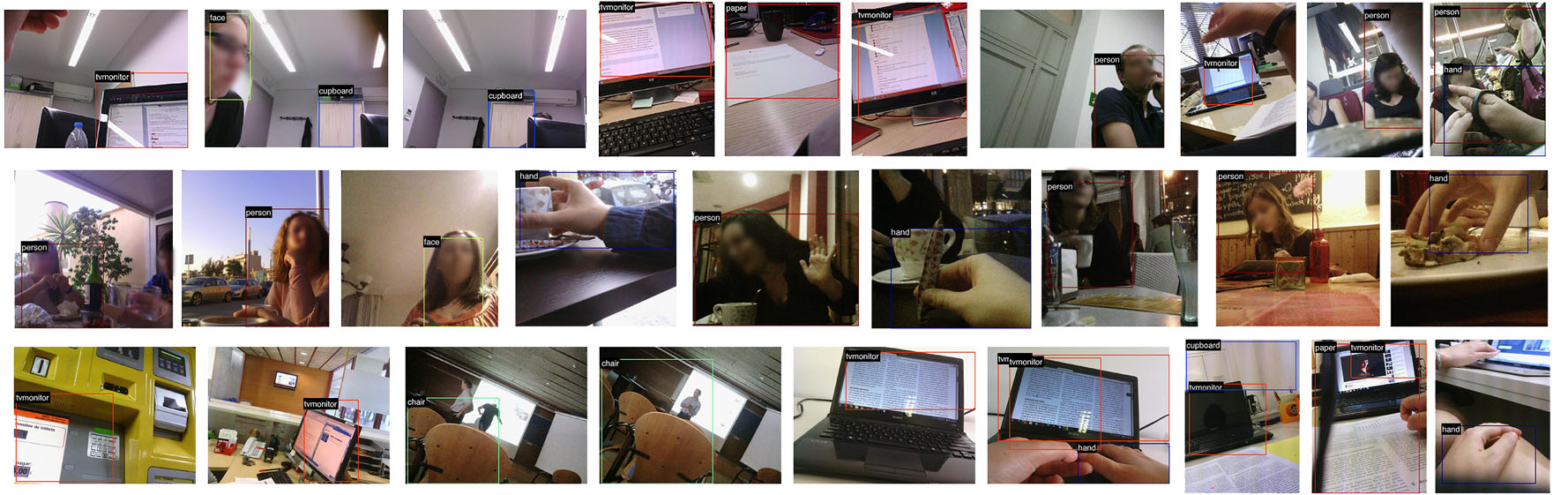} 
    \caption{Examples of discovered objects for three different subjects (one row each). Better viewed in digital format.}
    \label{fig:subjects_results}
\end{figure*}

%In figure \ref{fig:objects_distribution_maya} we show the relative \% of the distribution of objects found for all the datasets from one of our users.

%\begin{figure}[ht]
%    \centering
%	\includegraphics[width=\columnwidth]{objects_distribution_maya} 
%    \caption{Distribution of objects found for the two datasets of one of our users. Comparison for each method compared to the GT and the ones found by the Objectness measure (Ferrari).}
%    \label{fig:objects_distribution_maya}
%\end{figure}

\begin{figure}[ht]
    \centering
	\includegraphics[width=\columnwidth, height=4cm]{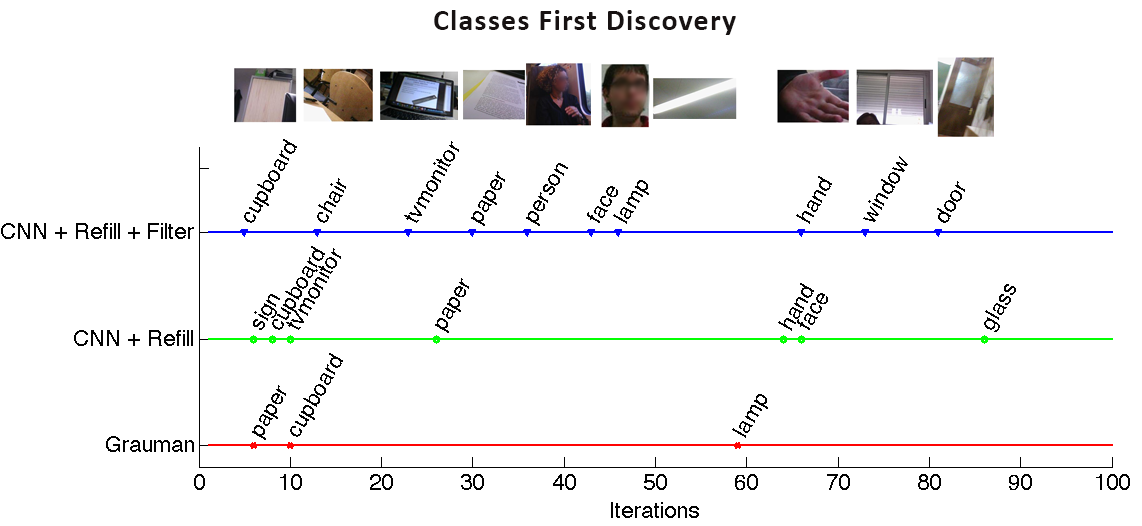} 
    \caption{First discovery of the object classes as a function of iterations.}
    \label{fig:classes_first_discovery}
\end{figure}

If we check the discovery order of the classes in each of the methods (see Fig. \ref{fig:classes_first_discovery}), we can see that some classes are more easily discovered and repeated over the following iterations than others. This is caused not only by the number of class instances appearing in the dataset, but also by the previously acquired knowledge (refill), the general method used, and/or the intra-class variability.

If we analyse the clusters number, where we find each class (see Table \ref{tab:num_clusters_classes}), we can see that even though having the same percentage of NO candidates (92.75\%), using Grauman's features (setting S1), we get 96\% of the clusters labeled as NO, but only 71\% of them using CNN + Refill (setting S3). Then, comparing it when adding the SVM filtering (setting S5), we can see that it gets reduced to a 49\% of the clusters thanks to the dramatic reduction of NO instances in the pool of unlabeled samples.

\begin{figure}[ht]
    \centering
	\includegraphics[width=\columnwidth]{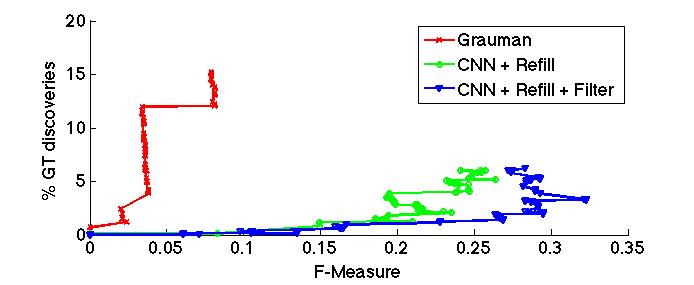} 
    \caption{Percentage of GT object discoveries accumulated on each iteration w.r.t. the F-Measure obtained.}
    \label{fig:GT_discoveries}
\end{figure}

In Fig. \ref{fig:GT_discoveries}, we can see the evolution of GT unique instances discovered by each of the methods on the accumulated iterations (each data point corresponds to an algorithm iteration) w.r.t. the F-Measure obtained by the method.

We can see that using Grauman's features seems to cover a wider variety of object samples than either with settings S3 or S5 (about 16\% against about 6-7\% of the GT samples). This result is probably directly related to the lower F-Measure obtained. Due to the lower generalization and representation capabilities of the set of features used (compared to CNN), the labeled clusters contain a wider variety of samples and objects, causing to label more unique object instances, but at the same time having a worse average result.

In Fig. \ref{fig:subjects_results} there are some examples of objects discovered by our methodology. We can see that it is able to discover instances of the same classes even having a high intra-class variability (person or hand). Note that some samples are not yet discovered due to the limited number of iterations applied (100).

Regarding the complexity of EOD, it is easy to see that (independently to the length of our feature vectors):
\begin{itemize}
    \item The objectness score extraction is of complexity $O(N)$, being $N$ the number of images in the dataset;
    \item The SVM filtering has complexity $O(N)$;
    \item The sorting of easiest objects is $O(N*W log(N*W))$, being $W$ the number of candidates extracted for each image;
    \item The refill strategy is $O(1)$; 
    \item The CNN features extraction is $O(M)$, being $M$ the easy objects number in the current iteration;
    \item The clustering of easy objects is $O(M^2)$;
    \item The best cluster labeling is $O(1)$;
    \item The one-class SVM cost is $O(M)$.
\end{itemize}
Leading in total a cost of $O(N*W log(N*W)+M^2)$, for each iteration.

%========================================================%
%========================================================%
%				CONCLUSIONS
%========================================================%
%========================================================%
\section{Conclusions} \label{sec:conclusions}

In this paper, we proposed a novel semi-supervised object discovery algorithm for egocentric data that relies on features extracted from a pre-trained CNN and uses a refill strategy for finding easily the classes with less samples. Moreover, we added a SVM filtering strategy for discarding a great part of the high amount of 'No Object' classes produced by any of the objectness measures. We compared 4 of the state of the art objectness measures in terms of 'No Object' instances produced and the Detection Rate obtained when extracting a low number of object candidates (W=50). We proved that the CNN features, the refill strategy (and the SVM filtering) can produce much better F-Measure results and can discover a larger number of infrequent classes than the state of the art approach on three datasets (MSRC, PASCAL 12 and EDUB), either being from general easy images, to egocentric and very difficult ones. Furthermore, we proved that this combined strategy also works better than the previous ones for very noisy and blurry images.

%========================================================%
%========================================================%
%				FUTURE WORK
%========================================================%
%========================================================%
\section{Future Work} \label{sec:future work}

Our future work involves the following tasks:

\begin{enumerate}
%    \item Improve the objectness measure or propose a new one trained using lifelogging images. % TOO DETAIL
    %\item %Extend our object discovery including a context-awareness term...
    \item Define an algorithm to discover objects, scenes and people to characterize the environment of the persons wearing the camera,
    %\item 
    \item Propose an iterative and combined scene and object discovery to take profit of the samples discovered from the complementary categories, and
    %\item 
    \item Make the method discriminative i.e. to detect which are the objects and scenes that characterize the environment of a person and distinguish them with respect to those of the other people.
%    \item Extend both the number of images and the number of subjects from our new EDUB public dataset. % TOO DETAIL
\end{enumerate}

\section*{Acknowledgments} 
This work was partially founded by the projects TIN2012-38187-C03-01 and SGR 1219.

\bibliographystyle{model2-names}
\bibliography{bib_file.bib}

\end{document}